\documentclass{article}
\usepackage{amsthm}
\usepackage{amssymb}
\usepackage{subfig}
\usepackage{graphicx}
\usepackage{tikz}
\usetikzlibrary{arrows}
\usepackage{pgfplots}
\usepackage{algorithm2e}
\usepackage{ifthen}
\usepackage{amsmath, amsthm, amsfonts}
\usepackage{aliascnt}

\newcommand{\etal}{et.al.}

\newboolean{showTodos}
\newboolean{showRemarks}
\newboolean{showExclude}

\setboolean{showTodos}{true}
\setboolean{showRemarks}{true}
\setboolean{showExclude}{false}

\newcommand{\todo}[1]{\ifthenelse{\boolean{showTodos}}{\optionalTextFormat{Todo}{#1}}{}}
\newcommand{\remark}[1]{\ifthenelse{\boolean{showRemarks}}{\optionalTextFormat{Remark}{#1}}{}}
\newcommand{\exclude}[1]{\ifthenelse{\boolean{showExclude}}{\optionalTextFormat{Exclude}{#1}}{}}

\newcommand{\alternative}[3]{
  \ifthenelse{\boolean{#1}}{
    [\textit{Start of Alternative}] #2
    
    \optionalTextFormat{Alternative Text}{#3}
  }{#2}
}

\newcommand{\optionalTextFormat}[2]{[\textbf{#1:~}\textit{#2}]}

\newcommand{\refeq}[1]{(\ref{#1})}

\newcommand{\feed}{\nonumber \\}

\theoremstyle{plain}
\newtheorem{theorem}{Proposition}

\newaliascnt{lemma}{theorem}
\newtheorem{lemma}[lemma]{Lemma}
\aliascntresetthe{lemma}

\newaliascnt{conj}{theorem}
\newtheorem{conjecture}[conj]{Conjecture}
\aliascntresetthe{conj}

\newtheorem*{corollary}{Corollary}

\newtheoremstyle{def}
{3pt}
{3pt}
{}
{\parindent}
{\itshape}
{:}
{.5em}
{}

\theoremstyle{definition}
\newtheorem{definition}{Definition}

\newcommand{\DeclareMathVarIndex}[2]{\DeclareMathOperator{#1}{#2}}

\renewcommand{\vec}[1]{\mathbf{#1}}
\newcommand{\mat}[1]{\mathbf{#1}}
\newcommand{\fun}[1]{\mathcal{#1}}
\newcommand{\set}[1]{\mathcal{#1}}

\DeclareMathOperator*{\argmin}{argmin}
\DeclareMathVarIndex{\opt}{*}

\renewcommand{\|}{\,|\,}

\newcommand{\Gaus}{\mathcal{N}}
\newcommand{\StdGaus}[3]{\Gaus\left(#1 | #2; #3\right)}

\newcommand{\Uniform}{\mathcal{U}}
\newcommand{\EExp}[2]{\left< #2 \right>_{#1}} 
\DeclareMathOperator{\opKL}{KL}

\newcommand{\KL}[2]{\opKL\left(#1 |\!| #2\right)}

\newcommand{\Int}[1]{\int_{#1}}

\renewcommand{\det}[1]{\left| #1 \right|}
\newcommand{\trans}[1]{#1^{T}}
\newcommand{\inv}[1]{#1^{-1}}

\DeclareMathVarIndex{\MAP}{MAP}
\newcommand{\Real}{\mathbb{R}}
\newcommand{\Integer}{\mathbb{N}}

\newcommand{\deltaFun}[1]{\delta_{#1}}

\newcommand{\abs}[1]{|#1|}

\newcommand{\F}{\ensuremath{\fun{F}}}
\newcommand{\Q}{\ensuremath{\fun{Q}}}
\newcommand{\xt}{\ensuremath{x_t}}
\newcommand{\yt}{\ensuremath{x_{t+1}}}
\newcommand{\ut}{\ensuremath{u_t}}
\newcommand{\rt}{\ensuremath{r_t}}
\newcommand{\xT}{\ensuremath{\bar{x}}}
\newcommand{\yT}{\ensuremath{x_{1\dots T}}}
\newcommand{\uT}{\ensuremath{\bar{u}}}
\newcommand{\rT}{\ensuremath{\bar{r}}}
\newcommand{\C}{\ensuremath{\fun{C}}}
\newcommand{\Ct}{\ensuremath{\fun{C}_t}}

\newcommand{\hCt}{\ensuremath{\hat{\fun{C}}_t}}

\newcommand{\nullpi}{\ensuremath{{\pi^{0}}}}
\newcommand{\optpi}{\ensuremath{{\pi^*}}}
\newcommand{\roptpi}{\ensuremath{{\bar{\pi}^*}}}
\newcommand{\ipi}{\ensuremath{{\pi^{i}}}}
\newcommand{\jpi}{\ensuremath{{\pi^{i+1}}}}
\newcommand{\bPsi}{\ensuremath{\bar{\Psi}}}
\newcommand{\LSP}{LS$\Psi$}
\newcommand{\AICOT}{AICO-T}

\usepackage{fullpage}
\usepackage{hyperref}
\usepackage{appendix}

\usepackage{authblk}
\title{Approximate Inference and Stochastic Optimal Control}
\author[1]{Konrad Rawlik}
\author[2]{Marc Toussaint}
\author[1]{Sethu Vijayakumar}
\affil[1]{Statistical Machine Learning and Motor Control Group, University of Edinburgh}
\affil[2]{Machine Learning and Robotics Group, TU Berlin}
\date{\today}

\begin{document}
\maketitle

\begin{abstract}
We propose a novel reformulation of the stochastic optimal control problem as an approximate inference problem, demonstrating, that such a interpretation leads to new practical methods for the original problem. In particular we characterise a novel class of iterative solutions to the stochastic optimal control problem based on a natural relaxation of the exact dual formulation. These theoretical insights are applied to the Reinforcement Learning problem where they lead to new model free, off policy methods for discrete and continuous problems. 
\end{abstract}
\tableofcontents
  \section{Introduction}
In recent years the framework of \emph{stochastic optimal control} (SOC) \cite{Stengel} has found increasing applicability in the domain of planning and control of realistic robotic systems \cite{Djordje10:ICRA, Toussaint10:ICRA} while also finding widespread use as one of the most successful normative models of human motion control \cite{Todorov02, Diedrichsen09}. In general SOC  can be summarised as the problem of controlling a stochastic system so as to minimise expected cost. The general problem subsumes a variety of different problems all based on slightly different assumptions, e.g. Markov Decision Processes \cite{Szepesvari09}, Reinforcement Learning \cite{SuttonBarto} or Adaptive Control. The increased use of the general formalism in high dimensional and non linear settings necessitates the development of novel efficient methods, while it's diverse nature makes novel theoretical insights into the general problem extremely desirable.

In the most general setting, the stochastic optimal control problem with arbitrary dynamics and cost function is analytically intractable and significant previous research has focused on developing efficient \emph{approximate} solution methods \cite{Jacobson70:DDP,Li06:iLQG}. In particular there have been, in recent years, an increasing number of attempts to relate the stochastic optimal control problem to problems from the domain of probabilist inference, specifically maximum likelihood problems, e.g., \cite{Toussaint06:EM, Barber09:EM}, and inference problems \cite{Kappen09:KL, Toussaint09:AICO}. The hope was that by finding such correspondences, the large number of available efficient Machine Learning \cite{Bishop} approaches will become applicable to the stochastic optimal control problem.

In this paper we propose a reformulation of the general stochastic optimal control problem as a problem of approximate probabilistic inference. Unlike previous theoretical work on this issue \cite{Kappen05:PI,Todorov09:PNAS,Kappen09:KL,Mitter03} this reformulation is exact without making further assumptions, though this comes at the cost of a lack of a closed form solution. However, the exact reformulation of stochastic optimal control as an inference problem is, in itself, not the main motivation of this work. Rather we see it as a starting point for development of novel approaches to the problem, which draw from the alternative interpretation. We show for example that the reformulation can be directly related to the previously proposed approximate inference control framework \cite{Toussaint09:AICO} which allows us to clarify the relation of the latter to stochastic optimal control.

Importantly we demonstrate that a, in the context of a probabilistic interpretation, natural relaxation of the new formulation directly leads to a novel class of iterative solutions for the stochastic optimal control problem. We characterise the form of these iterations and highlight their relation to previous applications of Expectation Maximisation algorithm in this area \cite{Toussaint06:EM, Barber09:EM}. We also directly demonstrate the applicability of these results, by deriving novel model free, off policy Reinforcement Learning algorithms for discrete and continuous problems.

We would like to note that this text forms part of the first author's PhD progress report (May 2010) submitted to the University of Edinburgh Graduate School. This document aims to make this work available to a wider audience as we have been made aware of the recent work by Kappen \etal\ \cite{Kappen:DPP} (pursued independently and in parallel) which shows distinct parallels to the methods developed here, with specifically the \emph{Dynamic Policy Programming} (DPP) algorithm having significant overlap with the here proposed \LSP\ algorithm, although the motivation and derivation differ. Furthermore we claim that the results presented here go beyond the work of \cite{Kappen:DPP} by providing a more general framework, relating it to previous approaches in Stochastic Optimal Control and Reinforcement Learning, and by demonstarting applicability of the algorithm to continuous problems. In particular we highlight a class of approximations which lead to analytical expressions in the continuous setting, mitigating the need to use computationally expensive numerical or Monte Carlo methods anticipated by \cite{Kappen:DPP}.


The remainder of this paper is structured as follows. After introducing necessary concepts of stochastic optimal control in \autoref{sec:prelim} we present in \autoref{sec:theory} our theoretical results relating to the approximate
inference formulation of stochastic optimal control problems. These are then applied in \autoref{sec:rl} to the Reinforcement learning problem. 

  \section{Preliminaries}
\label{sec:prelim}
In the remainder of this text we will consider control problems which can be modeled by a \emph{Markov Decision Process} (MDP) and before proceeding we first recall the standard formalism. We shall keep this exposition rather brief, only introducing concepts necessary for the development of the theory and methods in this paper. For a broader review one may refer to the 1$^{\text{st}}$ Year proposal or \cite{Szepesvari09}, or for a more thorough treatment to any of the numerous text books on the subject, e.g., \cite{Stengel, SuttonBarto, Bertsekas95}.

A MDP provides in general a model of a sequential decision process, where an agent observes it's state, chooses a control and then transitions to a new state whist incurring a certain cost. More formally, let $\xt \in \set{X}$ be the state and $\ut \in \set{U}$ the control signals at times $t = 1,2,\dots,T$. In order to simplify notation we will denote whole state and control trajectories $x_{1\dots T}$,$u_{0\dots T}$ by $\xT$,$\uT$. Let $P(\yt|\xt,\ut)$ be the transition probability for moving from $\xt$ to $\yt$ under control $\ut$ and let $\Ct(x,u) \geq 0$ be the cost incurred for choosing control $u$ in state $x$ at time $t$. A policy for time step $t$, $\pi_t(\ut|\xt)$, is the conditional probability of choosing the control $\ut$ given the state $\xt$. In the interrest of a less cluttered notation we shall in the following in general drop the subscript $t$ on $\pi$ if it is obvious from the context. An important family of policies is the set $\set{D}$ of \emph{deterministic policies}, which are policies given by a conditional delta distribution, i.e. $\pi(\ut|\xt) = \deltaFun{\ut = \tau(\xt)}$ for some function $\tau$. The stochastic optimal control problem consists of finding a deterministic policy\footnote{n.b. it can be shown that for problems of the type described here there exists a optimal policy which is deterministic \cite{Szepesvari09}} which minimises the expected cost, i.e., solving 
\begin{equation}
  \pi^{\opt} = \argmin_{\pi \in \set{D}} \EExp{q_\pi}{\sum_{t=0}^T\Ct(\xt,\ut)}~,
\end{equation}
where 
\begin{equation}
  \label{eq:marginal}
    q_\pi(\xT,\uT|x_0) = \pi(u_0|x_0)\prod_{t=1}^{T}\pi(\ut|\xt)P(\yt|\xt,\ut) ~,
\end{equation}
is the distribution over trajectories with start state $x_0$ and under policy $\pi$.

In the case of an infinite time horizon, i.e. for $T\rightarrow\infty$, we will restrict ourselves to the discounted cost formulation. That is we will assume the cost to be a discounted time stationary cost, so that $\Ct(\xt,\ut) = \gamma^t\C(\xt,\ut)$ for some discount factor $\gamma \in [0,1]$.

For a given policy $\pi$ we may define the value function $\fun{J}_t^{\pi} : \set{X} \rightarrow \Real$, as the mapping from a state $x$ to the expected cost of starting in $x$ at time $t$ and following $\pi$ thereafter, i.e.,
\begin{equation}
   \fun{J}_t^{\pi}(x) = \EExp{q_\pi(x_{t+1\dots T}, u_{t\dots T}|\xt = x)}{\sum_{k = t}^{T} \Ct(\xt,\ut)} ~.
\end{equation}
Similarly we may, for a given policy $\pi$, define the state-control, or state-action as it is more commonly known, value function $\Q^{\pi}_t : \set{X} \times \set{U} \rightarrow \Real$, which for a given $x,u$ gives the expected cost of starting in state $x$ at time $t$, choosing control $u$ and following $\pi$ thereafter, i.e.,
\begin{equation}
  \label{eq:Qfunction}
  \Q^{\pi}_t(x,u) = \EExp{q_\pi(x_{t+1\dots T}, u_{t+1\dots T}|\xt = x,\ut = u)}{\sum_{k = t}^{T} \Ct(\xt,\ut)} ~.
\end{equation}
Of obvious interest are the value and state action value functions of $\optpi$, which we denote by $\fun{J}^*$, $\fun{Q}^*$. They are sufficient, in the case of $\fun{J}^*$ together with knowledge of the transition probability and cost function, to characterise the optimal policy. An equation of particular importance in this context is the Bellman optimality equation
\begin{equation}
\label{eq:bellman:finite}
  \fun{J}^*_t(\xt) = \min_{\ut\in\set{U}} \left[\Ct(\xt,\ut) + \int_{\yt} P(\yt\|\xt,\ut)\fun{J}^*_{t+1}(\yt)\right] ~,
\end{equation}
which in the infinite horizon discounted cost setting gives the following fixed point equation for the optimal value function,
\begin{equation}
    \label{eq:bellman:infinit}
    \fun{J}^*(x) = \min_{u\in\set{U}} \left[\C(x,u) + \gamma\int_y P(y\|x,u)\fun{J}^*(y)\right] ~.
\end{equation}
Although the Bellman equations are in general not analytically tractable, in either the finite or infinite horizon case, they have provided the starting point for a large number of approaches for solving the stochastic optimal control problem, and will indeed be closely related to the starting point of the formulation proposed in this paper.

As an aside we note that throughout this paper we will in general be working under the more general assumption of infinite control and state spaces and hence use integrals, as has been already done in the Bellman equations. This is done with the understanding that for discrete problems these simply reduce to finite sums.

\section{On Stochastic Optimal Control and KL divergences}
\label{sec:theory}
\suppressfloats[t] 
\suppressfloats[b]
We will now state our main theoretical results which will form the basis of the work presented in \autoref{sec:rl} and proposed future work. Specifically we will show how stochastic optimal control can be formulated as a approximate inference problem in a certain probablistic model. For the purpose of this paper, we define \emph{approximate inference}, as the approximation of a true posterior within some family of distributions by minimization of some divergence measure. The divergence measure which we will consider here is the Kullback-Leibler divergence, which, for two distributions $q$ \& $p$ over $\set{X}$, is defined as 
\begin{equation}
    \KL{q}{p} = \int_\set{X} q(x) \log \frac{q(x)}{p(x)} ~.
\end{equation}
After introducing the probablistic model in \autoref{sec:model}, we will, in \autoref{sec:dual}, state and discuss our general duality result. As this result does not directly lead to a closed form solution of the stochastic optimal control problem we will then proceed do demonstrate in \autoref{sec:iterative} that under a relaxation of the exact dual a novel class of iterative approaches arises, which allows for closed form iterations. We will then derive such iterations for the finite and infinite horizon case. Finally we will discuss the relations of these results to previous work in the field.

\subsection{Bayesian Model of Control Problems}
\label{sec:model}
In most general terms, we would define inference based control in terms of a Dynamic Bayesian Network which includes multiple state, task, and control variables in each time slice. We would distinguish three types of random variables,  state and control variables, defined as in the stochastic optimal control framework, and additionally a set of variables, which we will refer to as \emph{task variables}, which capture the achievement of the objective described by the cost. In general the states and controls are latent variables and we wish to marginalise the states and infer the controls. The task variables on the other hand are observed, in the sense that we aim to make inference about the controls in the case of an achieved task.

\begin{figure}[t] 
\centering
\begin{tikzpicture}[
  >=stealth,
  nodes={minimum size=5mm},
  continous/.style={circle,draw=black,thick}, 
  discreet/.style={circle,draw=black,thick}, 
  observed/.style={fill=black!10},
  font = \tiny,
  normal line/.style={-stealth},
  back line/.style={normal line},
  front line/.style={normal line,
           preaction={draw=white, -, 
           line width=3pt}},
]
\matrix[row sep=2.5mm, column sep=5mm] {
    \node (u0) [continous] {$u_0$}; 
  & \node (u1) [continous] {$u_1$}; 
  & \node (u2) [continous] {$u_2$}; 
  & \node (uK) {};
  &;\\

    \node (x0) [continous] {$x_0$}; 
  & \node (x1) [continous] {$x_1$}; 
  & \node (x2) [continous] {$x_2$}; 
  & \node (x_) {$\ldots$};
  & \node (xK) [continous] {$x_T$};\\
    \node (r0) [discreet,observed]  {$r_0$};
  & \node (r1) [discreet,observed]  {$r_1$};
  & \node (r2) [discreet,observed]  {$r_2$}; 
  &
  & \node (rK) [discreet,observed]  {$r_T$};\\
};

\path (u0) edge [->, bend right=50, back line] (r0)
      (u1) edge [->, bend right=50, back line] (r1)
      (u2) edge [->, bend right=50, back line] (r2);
\path (x0) edge [->, front line] (x1)
      (x1) edge [->, front line] (x2)
      (x2) edge [- , front line] (x_)
      (x_) edge [->, front line] (xK);
\path (u0) edge [->, front line] (x1)
      (u1) edge [->, front line] (x2)
      (u2) edge [- , front line] (x_)
      (uK) edge [->, front line] (xK);
\path (x0) edge [->, front line] (r0)
      (x1) edge [->, front line] (r1)
      (x2) edge [->, front line] (r2)
      (xK) edge [->, front line] (rK);
\end{tikzpicture}
\caption{\label{fig:model:base} The graphical model of for the Baysian formulation of the control problem in the finite horizon case. In the infinite horizon case we obtain a stochastic markov process.}
\end{figure}
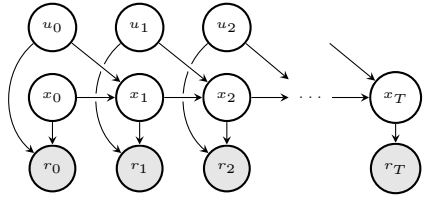 
More formally the model takes the form illustarted by the graphical model in figure \autoref{fig:model:base}. We relate the task likelihood to the classical cost by choosing
\begin{equation}
\label{eq:taskliklihood}
  P(\rt = 1 \| \xt, \ut) = \exp\{-\Ct(\xt,\ut)\}~,
\end{equation}
which is well defined due to the restriction $\Ct(\cdot,\cdot) \geq 0$. The complete joint is now given by
\begin{equation}
\label{eq:joint}
  P(\xT,\uT,\rT\|x_0; \pi) = q_{\pi}(\xT,\uT)\prod_{t=0}^T P(r_t\|u_t,x_t) ~,
\end{equation}
where $q_\pi$, the trajectory distribution under a policy, has been defined previously in \refeq{eq:marginal}. As indicated, our main interrest will be with the posterior under the \emph{assumed} observation of task success, and we will use the notation
\begin{equation}
\label{eq:posterior}
    p_\pi(\xT, \uT \| x_0) = P(\xT,\uT \| x_0, \rT = 1; \pi) = \frac{1}{P(\rT = 1 | x_0 ; \pi )}q_\pi(\xT,\uT)\prod_{t=0}^T P(r_t = 1 \|u_t,x_t)~.
\end{equation}

We would like to note that we have presented a model of sufficient structure for the results which follow, with the understanding that in cases with additional structure, better algorithm and stronger results may be obtainable. In particular we make no further assumption about conditional independence structure which is often present between subsets of state and control variables. Furthermore we view a single task variable in each time step as a sufficient representative of any set of task variables one might conceive, e.g., endeffector targets variables, collisions etc.. Eventually, the observed task variables only induce extra potentials on the remaining state and control variables. Therefore one could avoid introducing them at all in the formalism. However, we find their notion helpful to develop the theory.

\subsection{General Duality}
\label{sec:dual}
We may now directly state the main result relating the discussed Bayesian model to stochastic optimal control.
\begin{theorem}[Rawlik \& Toussaint \cite{Rawlik10:JMLR}]
  \label{thm:dual}
  Let $\nullpi$ be an arbitrary stochastic policy and $\set{D}$ the set of deterministic policies, then the problem
  \begin{equation}
      \label{eq:dual:KL}
      \argmin_{\pi \in \set{D}} \KL{q_\pi}{p_\nullpi} 
  \end{equation}
  is equivalent to the stochastic optimal control problem with cost per stage
  \begin{equation}
    \label{eq:dual:cost}
    \hCt(\xt,\ut) = \Ct(\xt,\ut) - \log \nullpi(\ut|\xt) 
  \end{equation}
\end{theorem}
\begin{proof}
  Let $\pi_t(u_t\|x_t) = \deltaFun{u_t = \tau_t(x_t)}$, for some function $\tau$, then
  \begin{align}
  \label{eq:dual:KLposterior}
    \KL{q_\pi}{p_\nullpi}
      & = \log P(\rT = 1)
        + \Int{\xT} \Int{\uT} q_{\pi}(\xT,\uT) \log \frac{q_{\pi}(\xT,\uT)}{q_{\nullpi}(\xT,\uT)}\feed
      & \qquad + \Int{\xT} \Int{\uT} q_{\pi}(\xT)~ \pi(\uT | \xT) \sum_{t=0}^T \log \frac{1}{\exp\{-\Ct(x_t, u_t)\}}  \\
      & = \log P(\rT = 1 | x_0; \nullpi)
        + \KL{q_{\pi}(\xT,\uT)}{q_{\nullpi}(\xT, \uT)}
        + \Int{\xT} \Int{\uT} q_{\pi}(\xT)~ \deltaFun{\uT = \tau(\xT)} \sum_{t=0}^T \Ct(x_t, u_t) \\
      & = \log P(\rT = 1 | x_0; \nullpi)
        + \KL{q_{\pi}(\xT,\uT)}{q_{\nullpi}(\xT, \uT)}
        + \Int{\xT} q_{\pi}(\xT)~ \sum_{t=0}^T \Ct(x_t, \tau_t(x_t))~.
   \end{align}
   Furthermore the divergence between the controlled process, $q_{\pi}$, and prior process, $q_{\nullpi}$ is
   \begin{align}
   \label{eq:dual:KLprior}
      \KL{q_\pi(\xT,\uT)}{p_\nullpi(\xT, \uT)}
        & = \Int{\xT} \Int{\uT} q_\pi(\xT,\uT) \sum_{t=0}^T \log \frac{\deltaFun{u_t=\tau_t(x_t)}}{\nullpi(u_t | x_t)} \\
        & = - \Int{\xT} q_\pi(\xT) \sum_{t=0}^T \log \nullpi(\tau_t(x_t) \| x_t) ~.
    \end{align}
    Hence, 
    \begin{equation}
      \KL{q_\pi}{p_\nullpi} = \log P(\rT = 1 | x_0; \nullpi) + \EExp{q_\pi}{\sum_{t=0}^T \left[\Ct(x_t, \tau_t(x_t)) - \nullpi(\tau_t(x_t) \| x_t) \right]} ~,
    \end{equation}
    and as $\log P(\rT = 1 | x_0; \nullpi)$ is constant w.r.t. $\pi$, the result follows.
\end{proof}
As an immediate consequence we obtain the direct equivalent for a given stochastic optimal control problem by,
\begin{corollary}
  With $\nullpi(\cdot|x) = \Uniform(\cdot)$, where $\Uniform(\cdot)$ is the uniform distribution over $\set{U}$, the problem in \refeq{eq:dual:KL} is equivalent to the stochastic optimal control problem.
\end{corollary}

One should note, that in general the result requires the set of controls to be such as to allow a uniform distribution to be defined, i.e., either finite or bounded. This is however merely a theoretical consideration, and although we will formally limit ourselves to cases where it is satisfied, it is of little practical consequence.

In general the presented reformulation of the stochastic optimal control problem will remain as intractable as the original formulation. In particular we can see that under the conditions of the corollary the $\opKL$ divergence reduces directly to the Bellman equation \refeq{eq:bellman:finite} plus a constant. This is a consequence of the fact that minimizing $\KL{q}{p}$, whilst restricting $q$ to be a delta distribution is equivalent to finding the maximum of $p$. Despite this intractability in the general case we argue that the presented formulation constitutes an interresting starting point for novel approaches to the problem. Both exact, iterative ones, as illustarted in the following section, but also approximate ones, as is the case with the Approximate Inference Control framework of Toussaint \cite{Toussaint09:AICO, Rawlik10:JMLR}, to which we relate this result in \autoref{sec:relation:AICO}.

\subsection{Iterative Solution}
\label{sec:iterative}
From the Bayesian point of view the restriction to delta distributions in \autoref{thm:dual} seems rather unnatural and can, as mentioned previously, be seen as the main cause why the $\opKL$ divergence remains intractable. A relaxation of this restriction, i.e. minimising w.r.t. to an arbitrary distribution $\pi(\cdot|\xt)$, makes as we shall show, the minimization tractable and although it obviously does not lead directly to a optimal policy, we have the following result 
\begin{theorem}
\label{thm:step}
For any $\pi \neq \nullpi$, $\KL{q_{\pi}}{p_{\nullpi}} \leq \KL{q_{\nullpi}}{p_{\nullpi}}$ implies
$\EExp{q_\pi}{\C(\xT,\uT)} < \EExp{q_\nullpi}{\C(\xT,\uT)}$.
\end{theorem}
\begin{proof}
Expanding the $\opKL$ divergences we have
\begin{multline}
    \KL{q_\pi(\xT,\uT)}{q_\nullpi(\xT,\uT)} + \EExp{q_\pi(\xT,\uT)}{\log P(\rt = 1|\xT,\uT)} + \log P(\rT = 1 \| x_0 ; \nullpi) 
    \\ \leq  \KL{q_\nullpi(\xT,\uT)}{q_\nullpi(\xT,\uT)} + \EExp{q_\nullpi(\xT,\uT)}{\log P(\rT = 1|\xT,\uT)} + \log P(\rT = 1\| x_0 ; \nullpi) ~.
\end{multline}
Subtracting $\log P(\rT = 1 \| x_0 ; \nullpi)$ on both sides and noting that $\KL{q_\nullpi(\xT,\uT)}{q_\nullpi(\xT,\uT)} = 0$, we obtain
\begin{equation}
  \KL{q_\pi(\xT,\uT)}{q_\nullpi(\xT,\uT)} + \EExp{q_\pi(\xT,\uT)}{\log P(\rT = 1|\xT,\uT)} \leq \EExp{q_\nullpi(\xT,\uT)}{\log P(\rT = 1|\xT,\uT)} ~.
\end{equation}
Hence, as $\KL{q_\pi(\xT,\uT)}{q_\nullpi(\xT,\uT)} \geq 0$ with equality iff $\pi = \nullpi$, the result follows.
\end{proof}
As an immediate consequence, with some initial $\nullpi$, the iteration 
\begin{equation}
    \label{eq:iteration}
    \pi^{i+1} \leftarrow \argmin_{\pi} \KL{q_{\pi}}{p_{\pi^i}}~,
\end{equation}
with $\pi$ an arbitrary\footnote{n.b. formally certain assumptions have to be made to ensure the support of $q_\pi$  is a subset of the support of $p_{\pi^i}$} conditional distribution over $u$, gives rise to a chain of stochastic policies with ever decreasing expected costs. However we note that \autoref{thm:step} has rather weak conditions and we can generalise the iteration as follows\footnote{n.b. a more general formulation is possible, which does not require $\pi^i \in \set{P}^i$, however the presented formulation suffices for our purpose}
\begin{theorem}
  \label{thm:iteration}
    Let $\set{P}$ be the set over all (stochastic) policies, if $\set{P}^i \subseteq \set{P}$ s.t. $\pi^{i} \in \set{P}^i$ for all $i$, then the policies in the sequence generated by
    \begin{equation}
        \label{eq:iteration:general}
        \pi^{i+1} \leftarrow \argmin_{\pi \in \set{P}^i} \KL{q_\pi}{p_{\pi^i}}
    \end{equation}
    have non increasing expected costs.
\end{theorem}
\begin{proof}
    As $\pi^{i} \in \set{P}^i$, $\KL{q_{\pi^{i+1}}}{p_{\pi^i}} \leq \KL{q_{\pi^{i}}}{p_{\pi^i}}$ and hence either \autoref{thm:step} applies or $\pi^i = \pi^{i+1}$.
\end{proof}
Note that this formulation admits \refeq{eq:iteration} as a special case.  A further interresting case is what we will refer to as \emph{asynchronous updates}. These are updates of only one time step at each iteration in any particular order, i.e. choose a schedule of time steps $\hat{t}^0, \hat{t}^1, \dots$ and let  $\set{P}^i = \{\pi \in \set{P} : \forall t \neq \hat{t}^i, \pi_t = \pi^i_t\}$.

Naturally questions about the behaviour in the limit of iterations covered by \autoref{thm:iteration} arise. As the expected cost is, under the assumption $\Ct(\cdot) > 0$ (cf. \autoref{sec:prelim}), bounded from below, we have as an immediate consequence 
\begin{corollary}
  Any iteration of the form \refeq{eq:iteration:general} converges.
\end{corollary}
This obviously leaves open the more interresting question, if, under what conditions and in what sense the policy converges to an optimal policy. Although it would certainly be desirable to obtain a general answer to this question, currently we are concentrating on the specific cases of \refeq{eq:iteration} and asynchronous updates for which we suggest that under weak conditions convergence to an optimal policy occurs (see \autoref{conj:convergence} below).

We will now proceed by first deriving specific updates for the finite horizon case, subsequently extending these to the infinite horizon, discounted cost setting.

\subsubsection{Finite Horizon Case}
\label{sec:theory:finite}
As indicated previously the general minimization of iteration \refeq{eq:iteration} can be performed analytically and here we provide the required derivation for the finite horizon case. To obtain the solution we bring the $\opKL$ divergence into the recursive form
\begin{multline}
\label{eq:KLrecursive}
  \KL{q_{\pi^{i+1}}(\xT,\uT)}{p_{\pi^i}(\xT,\uT)} = \int_{u_0} \pi^{i+1}(u_0\|x_0) \left[\log \frac{\pi^{i+1}(u_0\|x_0)}{\pi^i(u_0\|x_0)P(r_0\|x_0,u_0)} \right. \\ \left.+ \int_{\hat{x}} P(\hat{x}\|x_0,u_0)\KL{q_{\pi^{i+1}}(x_{2:T},u_{1:T}|x_1 = \hat{x})}{p_{\pi^i}(x_{2:T},u_{1:T}|x_1 = \hat{x})}\right]
\end{multline}
and utilizing the following general result
\begin{lemma}
\label{thm:minimizer}
Let $a,b,c$ be random variables with joint $P(a,b,c) = P(a)P(b\|a)P(c\|b,a)$ and $\set{P}$ the set of distributions over $a$, then
\begin{equation}
\label{eq:minimizer}
P(a)\exp\{\int_{b}P(b\|a) \log P(c=\hat{c}\|b)\} \varpropto \argmin_{q\in\set{P}} \KL{q(a)P(b\|a)}{P(a,b\|c = \hat{c})}
\end{equation}
and 
\begin{equation}
\label{eq:minimum}
\int_a P(a)\exp\{\int_{b}P(b\|a) \log P(c=\hat{c}\|b)\} = \min_{q\in\set{P}} \KL{q(a)P(b\|a)}{P(a,b\|c = \hat{c})}~.
\end{equation}
\end{lemma}
\begin{proof}
  see Appendix \ref{appendix:lemmas}
\end{proof}
Specifically assume the minimised nested $\opKL$ divergence for some time step $t+1$ is given by some $\exp\{\bPsi_{t+1}(\yt)\}$. Using the recursive formulation \refeq{eq:KLrecursive} and applying \refeq{eq:minimizer} with $a = \ut\|\xt$, $b = \yt$ and $P(c = \hat{c}\|b) = \exp\{\bPsi_{t+1}(\yt)\}P(\rt|\xt,\ut)$, it is easy to see that the new policy is given by the Boltzmann like distribution,
\begin{equation} 
  \pi^{i+1}(\ut|\xt) = \exp\{\Psi^{i+1}_t(\xt, \ut) - \bPsi^{i+1}(\xt)\} ~,    
\end{equation}
with energy 
\begin{equation}
  \label{eq:psi_t}
  \Psi^{i+1}_t(\xt,\ut) = \log \pi^i(\ut\|\xt) + \log P(\rt = 1\|\xt,\ut) + \int_{\yt} P(\yt\|\xt,\ut)\bPsi^{i+1}_{t+1}(\yt)
\end{equation}
and log partition function 
\begin{equation}
  \bPsi_t(\xt) = \log\int_u\exp\{\Psi(\xt,u)\} ~.
\end{equation}

Thus we can obtain the result for iteration \refeq{eq:iteration} by applying \refeq{eq:psi_t}  backwards in time, with $\bPsi^{i+1}_{T+1} = 0$ as the base case. Similarly asynchronous updates can be obtained by applying \refeq{eq:psi_t} only at one time step.

We now turn to the question of the behaviour of these updates in the limit. Let us define the following restricted optimal policy $\roptpi$
\begin{definition}
  Let $\set{U}_t^0(x) \subseteq \set{U}$ be the support of $\pi_t^0(\cdot\|x)$ and let $\set{U}^*_t(x)$ be the optimal controls at time $t$ in state $x$. If $\bar{\set{U}}^*_t(x) = \set{U}^*_t(x)\bigcap\set{U}_t^0(x)$ is not empty, $\roptpi(\cdot | x)$ is defined as the uniform distribution over $\bar{\set{U}}^*_t(x)$.
\end{definition}
Although we do not have any formal results yet, we suggest the following preliminary conjecture which we aim to complete in the near future
\begin{conjecture}
  \label{conj:convergence}
    Under weak assumptions, for both \refeq{eq:iteration} and asynchronous updates,  
    \begin{itemize}
        \item $\pi^i$ converges weakly to $\roptpi$
        \item $\bPsi_t$ converges pointwise to $-\fun{J}_t^* + c_t$, with $\fun{J}_t^*$ the optimal value function and $c_t$ a constant
    \end{itemize}
\end{conjecture}

\subsubsection{Infinite Horizon Case}
We will now consider the discounted infinite horizon setting. We proceed rather informally, but aim in future to formalise this setting as a limit case of the finite horizon setting.

It is sufficient to only consider time stationary policies in this setting \cite{SuttonBarto}. Under such a policy the entire process is time stationary, and, with a slight abuse of notation, we have 
\begin{equation}
  q_{\pi}(x_{>1},u_{>0}\|x_0 = \hat{x}) = q_{\pi}(x_{>2},u_{>1}\|x_1 = \hat{x}) ~.    
\end{equation}
It is now easy to show that
\begin{multline}
\KL{q_{\pi^{i+1}}(x_{>2},u_{>1}\|x_1 =\hat{x})}{p_{\pi^i}(x_{>2},u_{>1}\|x_1=\hat{x})} =\\ \gamma\KL{q_{\pi^{i+1}}(\xT,\uT\|x_0 =\hat{x})}{p_{\pi^i}(\xT,\uT\|x_0 =\hat{x})} ~,
\end{multline} 
which leads to the time stationary analog of \refeq{eq:psi_t},
\begin{equation}
  \label{eq:psi}
  \Psi^{i+1}(x,u) = \log \pi^i(u\|x) + \log P(r = 1\|x,u) + \gamma\int_{y} P(y\|x,u)\bPsi^{i+1}(y) ~.
\end{equation}
However due to the form of $\bPsi^{i+1}$, this does not yield $\Psi^{i+1}$ directly. Therefore we propose, in analogy to value iteration, e.g., \cite{SuttonBarto}, the update  
\begin{equation}
  \label{eq:psi_iteration}
  \Psi^{i+1}(x,u) \leftarrow \Psi^i(x,u) - \bPsi^i(x) + \log P(r = 1|x,u) + \gamma\int_{x'} P(x'|x,u)\bPsi^{i}(x')~.
\end{equation}
which corresponds to the assumption that after one step the old policy $\pi^i = \exp\{\Psi^i(x,u) - \bPsi^i(x)\}$ is followed. Although this update does not correspond to the iteration of \refeq{eq:iteration}, it can be constructed from a specific schedule of asynchronous updates. Specifically consider the schedule given with $\hat{t}^{j,k}$, where for each $j = 1,2,\dots$ updates are performed at $k = j, j-1, j-2, \dots, 0$. It is easy to see that after each update $\hat{t}^{j,0}$, the first step policy equals $\pi_0^i$, the policy obtained from \refeq{eq:psi_iteration}. Hence as this iteration falls into the class of \autoref{thm:iteration} we immediately obtain the guarantee of non increasing expected costs and convergence. Furthermore we anticipate that the schedule will satisfy the weak conditions of \autoref{conj:convergence} and its convergence to an optimal policy will directly follow.

\subsection{Relation to Previous Work}
\label{sec:relation}
In the following we will relate the presented work in greater detail to three recent developments in the field. However we note that attempts to relate stochastic optimal control to inference have along history, in part motivated by the exact duality for the \emph{linear-quadratic-gaussian} (LQG) case discovered by Kalmann \cite{Stengel}. In general the idea of replacing costs, utilities
or rewards by an auxiliary binary random variable has a long history \cite{Cooper88,Shachter88,DayanHinton97}.
Shachter \& Peot \cite{ShachterPeot92} even mention work by Raiffa (1969) and von Neumann \& Morgenstern (1947) in this context. Although approaches have varied between using the interpretation of cost as energy, together with the typical identification of energy with negative log probability, as a has been done here, and choosing probabilities which are proportional to the reward or utility.

\subsubsection{Approximate Inference Control}
\label{sec:relation:AICO}
As the \emph{approximate inference control} (AICO) framework was discussed in detail in the $1^{st}$ Year Report we will refrain from a full description here. In suffices to recall that AICO is formulated within the model described in \autoref{sec:model} and aims to find an approximation to $p_\nullpi$ by a message passing approach similar to Expectation Propagation \cite{Minka01:Phd}. Although the original work \cite{Toussaint09:AICO} observed a close relation of the messages in the LQG case to the classical Riccatti \cite{Stengel} equations, no claims were made regarding stochastic optimality and it was suggested to choose the \emph{maximum a posteriori} (MAP) controls. With the results presented in \autoref{sec:dual} the relation of AICO to stochastic optimal control can now be clarified. Specifically a possible interpretation for AICO is to see it as finding an approximation to $p_\nullpi$, such as to make the $\opKL$ divergence of \autoref{thm:dual} tractable. However,  even under this interpretation we note that the result of the minimization of the KL divergence, even under a Gaussian approximation to $p_\nullpi$, are not the MAP control, rather one should solve the Ricatti equation arising from the approximation.

\subsubsection{Path Integral and KL control}
In recent years several groups were independently able to show that for a restricted class of stochastic optimal control problems the minimized Bellman equation \refeq{eq:bellman:finite} becomes linear and the problem admits a solution in closed form \cite{Mitter03, Kappen05:PI, Todorov09:PNAS, Kappen09:KL}. These linear Bellmann equations can be seen as a $\opKL$ divergence \cite{Kappen09:KL}, leading to a close relation to the formulation in \autoref{thm:dual}. We will demonstrate this close relation in the discrete time case, leaving the continuous time case for future consideration as we have not yet developed it in our framework.

Let us briefly recall the KL control framework of Kappen \etal~\cite{Kappen09:KL}, the alternative formulations of Todorov \cite{Todorov07:MDP, Todorov09:PNAS} being equivalent. Choose some free dynamics $\nu_0(\yt|\xt)$ and let the cost be given as 
\begin{equation}
  \label{eq:KLcost}
  \C(\xT) = \ell(\xT) + \sum \log\frac{\nu(\xT)}{\nu_0(\xT)} 
\end{equation}
where $\nu(\yt|\xt)$ is the controlled process under some policy. Then
\begin{equation}
  \label{eq:KLmain}
  \EExp{\nu}{\C(\xT)} = \KL{\nu(\xT)}{\nu_0(\xT)\exp\{-\ell(\xT)\}}
\end{equation}
which is minimised w.r.t. $\nu$ by 
\begin{equation}
  \label{eq:KLsolution}
  \nu(\yT|x_0) = \frac{1}{Z(x_0)}\exp\{-\ell(\yT)\}\nu_0(\yT|x_0)
\end{equation}
and one concludes that the optimal control is given by $\nu(\yt|\xt)$, where presumably the implied meaning is that  $\nu(\yt|\xt)$ is the trajectory distribution under the optimal policy.

Although \refeq{eq:KLsolution} gives a process which minimises \refeq{eq:KLmain}, it is not obvious how to compute actual controls from this process. Specifically when given a model of the dynamics, $P(\yt|\xt,\ut)$, and having chosen some $\nu_0$, a non trivial, yet implicitly made, assumption is that
\begin{equation}
  \label{eq:KLassume}
      \exists \pi, \quad\text{s.t.}\quad \nu(\yt|\xt) = \Int{\ut}{P(\yt|\xt,\ut)\pi(\ut|\xt)} ~.
\end{equation}
In fact in general such a $\pi$ will not exists. This is made very explicit for the discrete MDP case in \cite{Todorov07:MDP}, where it is acknowledged that the method is only applicable if the dynamics are fully controlable, i.e., $P(\yt\|\xt,\ut)$ can be brought into any arbitrary form by the controlls. Although in the same paper it is suggested that solutions to classical problems can be obtained by \emph{continuous embedding} of the discreet MDP, such an approach has several drawbacks. For one it requires solving a continuous problem even for cases which could have been otherwise represented in tabular form, but more importantly such an approach is obviously not applicable to problems which already have continuous state or action spaces. In the latter case Kappen \etal~claim (cf. section 4 of \cite{Kappen09:KL}) that the KL control approach is applicable if the problem is of the following form
\begin{equation}
\label{eq:Kappen:problem}
\begin{aligned}
    \yt & = \F(\xt) + \mat{B}(\xt)(\ut + \xi), \quad \xi \sim \Gaus(0,\mat{Q})~,\\
    \Ct(\xt,\ut) & =  \ell(\xt) + \trans{\ut}\mat{H}\ut ~,
\end{aligned}
\end{equation}
with $\F, \mat{B}$ and $\ell$ having arbitrary form, but $\mat{H}, \mat{Q}$ are such that $\inv{\mat{H}} \propto \mat{Q}$. We dispute this claim, showing that, in the discreet time case, \refeq{eq:KLassume} is not fullfilled and that correcting the problem leads an equivalent of \autoref{thm:dual}.

It will be sufficient to consider the simplest possible case of a one dimensional, one time step LQG problem. Let \begin{equation}
    P(\yt|\xt,\ut) = \StdGaus{\yt}{\xt + \ut}{\Sigma}
\end{equation}
and
\begin{equation}
  \Ct(\xt,\ut) = \xt R \xt + \ut \inv{\Sigma} \ut ~.
\end{equation}
The claim made by Kappen \etal\ is, that for $\nu_0 = P(\yt | \xt, \ut = 0)$, the $\opKL$ formulation is equivalent to the corresponding stochastic optimal control problem. Or more specifically that 
\begin{equation}
  \nu(x_1|x_0) 
    \varpropto P(x_1|x_0,u_0 = 0)\exp\{- x_1 R x_1\} 
    = \StdGaus{x_1}{x_0}{\inv{\Sigma} + R}
\end{equation}
gives the optimal controls, hence \refeq{eq:KLassume} should hold. In particular, as we know the LQG problem has a unique deterministic stochastic optimal control solution \cite{Stengel}, there should be a deterministic $\pi$ s.t. \refeq{eq:KLassume} holds. But notice that we can not influence the variance of $P(\yt|\xt,\ut)$ by specific choices of a deterministic $\pi$, hence \refeq{eq:KLassume} does not hold. Specifically $\nu$ is \emph{not} the trajectory distribution under the optimal policy. In fact there may not even be a stochastic policy s.t. \refeq{eq:KLassume} holds. Consider the case when the cost 'variance' $\inv{R}$ is smaller then the variance of the noise, $\Sigma$. Then $\nu(x_1|x_0 = 0)$ will have variance smaller then $\Sigma$. But even though with a stochastic policy the variance of the marginal process can increase, it can not decrease.

The question now arises what controls should we choose?  A principled choice would be to choose $\pi$ to minimise a $\opKL$ divergence. The first intuition would be to take
\begin{equation}
  \label{eq:KLmarginalpolicy}
  \argmin_\tau \KL{P(\yt|\xt,\ut = \tau(\xt))}{\nu(\yt|\xt)} ~.
\end{equation}
However noting that 
\begin{align}
  \nu(\yt|\xt)
    & = \frac{1}{Z(\xt)}\nu_0(\yt|\xt)Z(\yt) \\
    & = \frac{1}{Z(\xt)}\nu_0(\yt|\xt)\EExp{\nu_0(x_{k+1:K}|\yt)}{\exp\{-\C(x_{k+1:K})\}} ~,
\end{align}
the $\opKL$ divergence can be written as
\begin{equation}
  \KL{P(\yt|\xt,\ut = \tau(\xt))}{\nu_0(\yt|\xt)} + \EExp{P(\yt|\xt,\ut = \tau(\xt))}{\log Z(\yt)} - \log Z(\xt) ~.
\end{equation}
This is the correct expression for the expected cost, if $\log Z$ is the value function, however the latter is only the case if the normalized form of the $\opKL$ divergence in \refeq{eq:KLmain} becomes zero at the minimum. Here we are specifically assuming this not to be the case, implying this formulation does not lead to stochastic optimal controls and we are therefore compelled to take
\begin{equation}
  \label{eq:KLfull}
  \argmin_\pi \KL{q_\pi(\xT)}{\nu(\xT)} ~.
\end{equation}
This is very similar to the $\opKL$ divergence in \autoref{thm:dual}. In fact, under the conditions of the corollary to \autoref{thm:dual} and if the problem is of the form in \refeq{eq:Kappen:problem}, we can write
\begin{equation}
  p_\nullpi(\xT,\uT) = \nu(\xT)\prod\exp\{(\yt - \xt - \F(\xt)\inv{\Sigma}\ut -  \frac{1}{2}\ut\inv{H}\ut)\}
\end{equation}
and the $\opKL$ divergence of \autoref{thm:dual} can alternatively be written as 
\begin{equation}
  \KL{q_\pi}{p_{\nullpi}} = \KL{q_\pi(\xT)}{\nu(\xT)} - \EExp{q_\pi(\xT,\uT)}{\sum (\yt - \xt - f(\xt)\inv{\Sigma}\ut -  \frac{1}{2}\ut\inv{H}\ut)} ~.
\end{equation}
Furthermore as for a deterministic policy, i.e. $\pi(\ut | \xt) = \deltaFun{\ut = \tau(\xt)}$, 
\begin{equation}
  \EExp{q_\pi}{(\yt - \xt - f(\xt)} = \EExp{q_\pi}{\ut} = \tau(\xt) ~,
\end{equation}
we can see that the second term is zero under the condition $\inv{H} = 2\inv{\Sigma}$, i.e. under the conditions required by Kappen \etal, and \refeq{eq:KLfull} is equivalent to the formulation in \autoref{thm:dual}.

\subsubsection{Expectation Maximization Approaches}
Several suggestions for mapping the stochastic optimal control problem onto a maximum likelihood problem and using \emph{Expectation Maximization} (EM) have been recently made in the literature \cite{Toussaint06:EM, Barber09:EM}. Going further back the probability matching approach \cite{DayanHinton97, Sabes96} is also closely related to expectation maximization procedures.

As one may suspect when considering \refeq{eq:iteration} our approach has a close relation to the free energy view of EM \cite{Hinton:GEM, Bishop}. In this view, EM alternates between minimizing $\KL{q(z)}{P(z|y;\theta)}$ w.r.t. $q$, where $z$,$y$ are the latent and observed variables and $\theta$ the parameters, and maximizing the free energy, defined as 
\begin{equation}
  \fun{L}(q,\theta) = \int_z q(z)\log\frac{P(z,y;\theta)}{q(z)}
\end{equation}
w.r.t. $\theta$. In our case $z$ and $y$ correspond to $\xT,\uT$ and $\rT$, while $\theta$ corresponds to $\pi$. It is easy to see that \refeq{eq:iteration}, or \refeq{eq:iteration:general} for that matter, correspond to a generalized E-Step. The \emph{generalized} indicates that only a partial step is performed, i.e., we are only lowering, rather then minimizing, $\KL{q(z)}{p(z|y;\theta)}$ w.r.t. $q$. Furthermore the choice of $\jpi$ corresponds to a generalized M-Step, as 
\begin{align}
  \fun{L}(q_{\jpi},\pi = \jpi) 
    & = \int_{\xT,\uT} q_{\jpi} \log P(\rT=1|\xT,\uT) \\
    & \geq \int_{\xT,\uT} q_\jpi \log P(\rT=1|\xT,\uT) - \KL{q_\jpi}{q_\ipi}  \\
    & = \fun{L}(q_\jpi, \pi = \ipi) ~.
\end{align}
Hence we conclude that our method corresponds to an generalized EM algorithm. 

Although we have shown that one can interpret the proposed approach in terms of EM we emphasise that it differs significantly from the applications of EM in previous work. For one we note that it is not our aim to find the maximum likelihood policy and in fact as we are using a generalized E-Step we lose the guarantee of convergence to a local maximum of the likelihood. In general maximizing the marginal log likelihood, the objective of EM, would not be desirable in our model, as despite the fact that for a given state and control trajectory, the classical cost and the task likelihood are directly related by 
\begin{equation}
  \C(\xT,\uT) = -\log P(\rT = 1| \xT,\uT)~,
\end{equation}
no such direct equality relation for the marginal likelihood can be obtained. Although using Jensen's inequality we may obtain
\begin{equation}
  \EExp{q_\pi(\xT,\uT)}{\C(\xT,\uT)} \leq -\log P(\rT = 1) ~,
\end{equation}
this bound, which has also been previously observed in \cite{Toussaint09:AICO}, is not necessarily tight and hence the optimal stochastic optimal solution does not necessarily coincide with the maximum likelihood solution. 

A more fundamental difference is that we can avoid finding an explicit representation for $q_\pi$. In both the approaches of \cite{Toussaint06:EM} and \cite{Barber09:EM} calculating $q_\pi$ explicitly is a major computational step and presents a problem if these methods were to be applied to the continuous setting where $q_\pi$ may not be analytically tractable. 

Finally we anticipate that \autoref{conj:convergence} will hold, giving a guarantee of convergence to an optimal policy which other EM methods can not provide.

\subsection{Conclusion}
The contribution of this section is a novel interpretation of the stochastic optimal control problem as an approximate inference problem and the derivation of a iterative solution to the control problem based on this new interpretation. The proposed approach has also been shown to have interresting links to other current research directions in the field. In particular we deomnstarte that the approach can be understood to underlie both the approximate inference control framework and, in the time discrete setting, the KL control framework. This theoretical work is intended to  provides the foundation for the remainder of this paper and future work.
\section{Reinforcement Learning}
\label{sec:rl}
\suppressfloats[t] 
So far we have assumed the transition model and cost function are readily available. We now turn to the reinforcement learning setting \cite{Kaelbling96:RL, Szepesvari09, SuttonBarto}, where one aims to learn a good policy only given samples from the transition probability and associated incurred costs.

We will demonstrate how the theoretical results previously derived can be applied to such problems yielding algorithms which are both \emph{model free} and \emph{off policy}. Model free indicates that the algorithm does not construct an explicit representation of the transition probability and cost function but rather directly learns a representation of the optimal policy. Off policy on the other hand means that the optimal policy can be learnt from samples collected under a different, often sub-optimal, policy.

We will proceed by first deriving a tabular algorithm which is applicable for problems with small, finite, state and control spaces, before subsequently extending it to problems with continuous state and control spaces by using approximate parametric representations. Both algorithm are applied to classical problems in the field.

\subsection{Finite Problems}
Let us consider problems in the infinite horizon discounted cost setting and recall that the update function for $\Psi$ suggested in \autoref{sec:iterative} for this case was
\begin{equation}
   \Psi(x,u) \leftarrow \Psi(x,u) - \bPsi(x) + \log P(r = 1|x,u) + \gamma\int_{x'} P(x'|x,u)\bPsi(x') ~.
\end{equation}
For any given $x,u$ this update can be written as an expectation w.r.t. the transition probability $P(y|x,u)$, and hence may be approximated from a set of sampled transitions. In particular given a single sample $(x,u,\ell,y)$ of a transition from $x$ to $y$ under control $u$ incurring cost\footnote{n.b. we assume we observe the cost, i.e., $\ell = -\log P(r = 1|x,u)$. } $\ell$ we may perform the approximate update
\begin{equation}
 \label{eq:psi_learning}
 \Psi(x,u) \leftarrow \Psi(x,u) + \left[\gamma\bar{\Psi}(y) - \bar{\Psi}(x) - \ell\right] ~.
\end{equation}
Given a stream of samples $x_0, u_0, \ell_0, x_1, u_1, \ell_1, \dots$ we can then apply such an update for each tuple $(\xt, \ut, \ell_t, \yt)$ individually. Without a particular justification we furthermore can introduce a decaying learning rate parameter, similar to other reinforcement learning algorithms, in order to damp these updates. In practise however we did not find such a learning rate to improve results significantly. We call the resulting algorithm \emph{$\Psi$-learning}. As indicated previously it is model free and can be employed for off policy learning.

\subsubsection{Relation to Classical Algorithms}
Before proceeding let us highlight certain similarities and differences between $\Psi$-learning and two classical algorithms, $\fun{Q}$-learning and TD(0) \cite{SuttonBarto}. 

As the name indicates, $\fun{Q}$-learning learns the state-action value function (cf. Equation \refeq{eq:Qfunction}). We note that $\Psi$ has certain similarities to a $\fun{Q}$ function, in the sense that a higher value of $\Psi$ for a certain control in a given state indicates that the control is \emph{'better'}. In fact for the optimal controls the $\fun{Q}$ function and $\Psi$ converge to the same value\footnote{n.b. at the moment this is conjecture, as it is a consequence of \autoref{conj:convergence}}. However unlike the $\fun{Q}$ function, which also converges to the expected cost for the sub-optimal controls, $\Psi$ goes to $-\infty$ for sub-optimal actions. A potentially more insightfull difference between the two algorithm is nature of updates employed. The $\fun{Q}$-learning algorithms uses updates of the form 
\begin{equation}
    \fun{Q}(x,u) \leftarrow \fun{Q}(x,u) + \alpha\left[\ell + \gamma\max_{u'}\fun{Q}(y,u') - \fun{Q}(x,u)\right] ~,
\end{equation}
where $\alpha$ is a learning rate. Note that it will employ only information from one current control and the best, according to current knowledge, future control. The $\Psi$-learning algorithm on the other hand uses $\bPsi$ which in some sense averages over information about the future according the current belief about the control distribution, rather then using single $\Psi$ values.

A connection to the TD(0) algorithm which learns a value function is given by the form of the update. The TD(0) update has the form
\begin{equation}
    \fun{J}(x) = \fun{J}(x) + \alpha\left[\ell + \gamma\fun{J}(y) - \fun{J}(x)\right]
\end{equation}
with $\alpha$ again a learning rate. We observe that as by \autoref{conj:convergence}, $\bPsi$ converges, up to a additive constant, to the value function of the optimal policy, the $\Psi$-learning update converges towards the TD(0) update for samples generated under the optimal policy. The emphasise is on, \emph{convergence} to the TD(0) update, in general it will not correspond to an TD(0) update. In particular a important differences between the two algorithms is that TD(0) is a on-policy method, that is it learns the value function of the policy used to generate samples, while the proposed $\Psi$-learning is off-policy.

\subsubsection{Results}
Problems with finite state and action spaces allow $\Psi$ to be represented directly in tabular form. We evaluated such a tabular $\Psi$-learning algorithm on the grid world domain \cite{SuttonBarto}. Specifically we used the following task formulation. The state space is given by a $N\times N$ grid with some states occupied by obstacles. The controls allow the agent to transition to any neighbouring state not occupied by an obstacle or to remain at the current state. A transition to a neighbouring state succeeds with probability 0.8, with the agent remaining at the current location in case of failure. Choosing to remain in the current state succeeds with probability 1. Additionally a set $\set{A}\subseteq\set{X}$ of absorbing target states, i.e., $1 = P(\yt \in \set{A}|\xt \in \set{A}, u \in \set{U})$, is defined. In every time step a cost of 1 is incurred if the agent is in any state which is not a target state, while at a target state no cost is incurred, i.e., $C(x,u) = \deltaFun{x \notin \set{A}}$ with $\deltaFun{}$ the Kronecker delta. The cost was not discounted, i.e., $\gamma = 1$.

We used tabular $\Q$-learning, e.g., \cite{SuttonBarto}, as a baseline. Both algorithms were run with controls sampled from an uninformed policy, i.e. a uniform distribution over the controls available at a state. Once a target state was reached, or if the target wasn't reached within 100 steps, the state was reset randomly. The learning rate for $Q$-learning decayed as $\alpha = c/(c + t)$ with $t$ the number of transitions sampled and $c$ a constant which was optimised manually.

\begin{figure} 
\centering
\subfloat[]{
\includegraphics[width=0.4\textwidth]{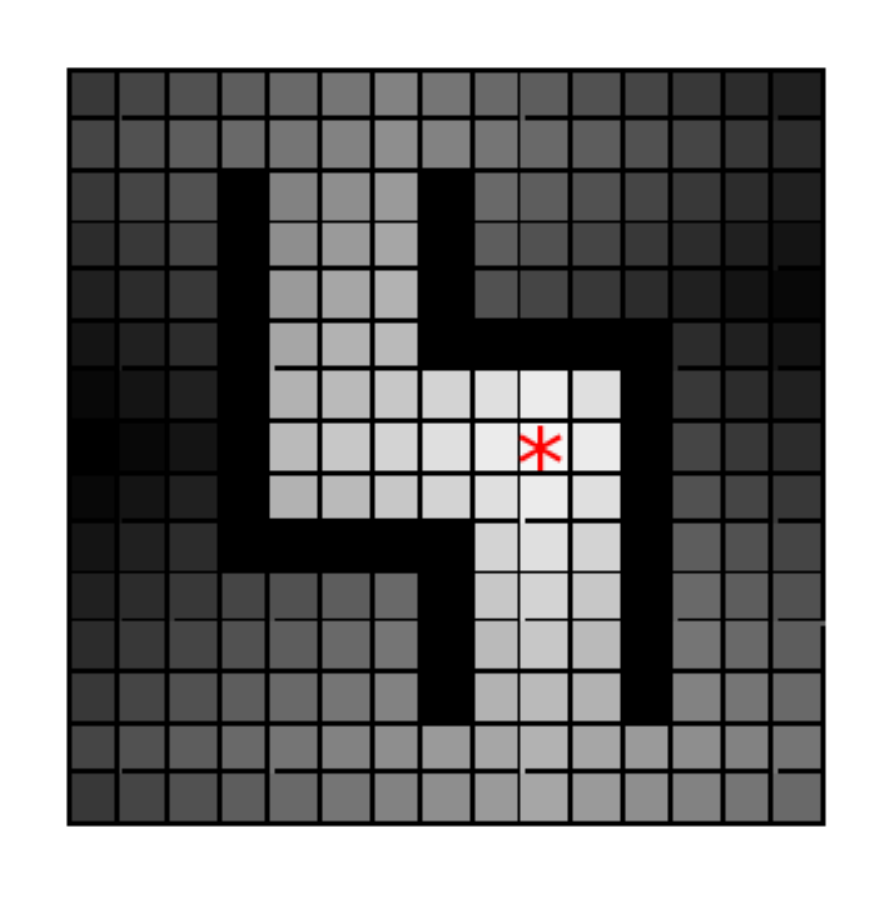}
}
\subfloat[]{
\input{figure/img_Q_Tod_Error}
}
\caption{\label{fig:Qlearning} Results for tabular $\Psi$-learning on an example grid world problem. \textbf{(a)} the optimal value function (white low expected cost - black high expected cost) of the problem. Obstacles are black and the target state is indicated by *. \textbf{(b)} Evolution of the mean error in \refeq{eq:Verr} averaged over 10 trials for each of the algorithms. Error bars indicate the standard deviation.}
\end{figure} 
Representative results for a single instance of the general task are given in figure \ref{fig:Qlearning}. We report the approximation error 
\begin{equation}
  \label{eq:Verr}
  e_{\fun{J}} = \frac{\max_x \abs{\fun{J}(x) - \hat{\fun{J}}(x)}}{\max_x \fun{J}(x)}    
\end{equation}
between the true value function $\fun{J}$, obtained by value iteration, and its estimate $\hat{J}$, given by $\bar{\Psi}$ and $\max_u Q(x,u)$ respectively. Both algorithm achieved the same error at convergence. However $\Psi$-learning consistently outperformed $Q$-learning in terms of the number of samples required to convergence. We additionally considered a greedy variant of $\Psi$-learning where the controls are sampled from the policy given by the current $\Psi$, i.e. $\pi(u|x) = \exp\{\Psi(x,u) - \bar{\Psi}(x)\}$. As expected we found that the greedy version greatly outperformed sampling using an uninformed policy.

\subsection{Continuous problems}
For continuous control problems, i.e. those with infinite state or controls sets, storing $\Psi$ in tabular form  clearly becomes impossible and even for discreet problems it may be impracticable due to the size of the table required. In such cases, it is common to resort to parametric representations \cite{Szepesvari09}, and here we follow such an approach to extend $\Psi$-learning to continuous problems. Although we will concentrate on the continuous case, we note that the proposed approach could also be employed for large discreet problems.

\subsubsection{The \LSP\ algorithm}
Similar to numerous previous approaches \cite{Boyan02, Peters03, SuttonBarto, Szepesvari09} we used a linear basis function model, to approximate $\Psi$, i.e., 
\begin{equation}
    \Psi(x,u) \approx \tilde{\Psi}(x,u,\vec{w}) = \sum_{i = 0}^M w_i\phi(x,u)\,
\end{equation}
where $\phi_i:~\set{X}\times\set{U}~\rightarrow~\Real$ are a set of given basis functions and $\vec{w} = (w_{1},\dots,w_M)$ is the vector of parameters we learn. For such an approximation and given set of samples $(x_{1\dots K},u_{1\dots K},\ell_{1\dots K},y_{1\dots K})$, the $\Psi$-learning update \refeq{eq:psi_learning} can be written in matrix notation as
\begin{equation}
    \Phi\vec{w}^{i+1} = \Phi\vec{w}^{i} + \vec{z}~,
\end{equation}
where $\Phi$ is the $K \times M$ matrix with entries $\Phi_{i,j} = \phi_i(x_j,u_j)$ and $\vec{z}$ is the vector with elements 
\begin{equation}
  \vec{z}_k = \gamma\bar{\Psi}(y_k) - \ell_k - \bar{\Psi}(x_k) ~.    
\end{equation}
From this we can obtain
\begin{equation}
  \vec{w}^{i+1} - \vec{w}^i = \inv{(\trans{\Phi}\Phi)}\trans{\Phi}\vec{z} ~, 
\end{equation}
which suggests the update rule
\begin{equation}
    \vec{w} \leftarrow \vec{w} + \inv{(\trans{\Phi}\Phi)}\trans{\Phi}\vec{z} ~.
\end{equation}
This is equivalent to computing the $\Psi$-learning update of \refeq{eq:psi_learning} for the current approximation and projecting the result onto the space spanned by the basis functions in the least squares sense, an approach which has seen repeated use in reinforcement learning \cite{Lagoudakis03, Szepesvari09}. We call the algorithm resulting algorithm, which, as tabular $\Psi$-learning, is a model free, off-policy method, \emph{Least Squares $\Psi$-learning} (LS$\Psi$).

The choice of basis functions for LS$\Psi$ is somewhat complicated by the need to evaluate the log partition function of the policy $\bar{\Psi}$, i.e. $\log\int_u \exp\{\tilde{\Psi}(x,u)\}$, when forming the vector $\vec{z}$. In cases where $\set{U}$ is a finite set, arbitrary basis functions can be chosen as the integral reduces to a finite sum. However for problems with infinite control spaces one needs to ensure the bases are chosen such that the arising integral is analytical tractable, i.e. the partition function of the stochastic policy can be evaluated. One class of basis sets for which this is the case, are those for which $\tilde{\Psi}(x,u,\vec{w})$ has the form
\begin{equation}
  \label{eq:approx_form}
  \tilde{\Psi}(x,u,\vec{w}) = -\frac{1}{2}\trans{u}\mat{K}(x,\vec{w})u + \trans{u}\vec{k}(x,\vec{w}) + k(x,\vec{w})
\end{equation}
where $\mat{K}(x,\vec{w})$ is a positive definite matrix. For such a set the integral is of the Gaussian form and the closed form solution
\begin{equation}
    \log\int_u \exp\{\tilde{\Psi}\} = - \log\det{\mat{K}} - \frac{1}{2}\vec{k}'\inv{\mat{K}}\vec{k} + k + constant
\end{equation}
is obtained. Obviously the implication of such a basis set is that the policies are restricted to conditional Gaussian distributions. Specifically the policy is given by 
\begin{equation}
  \pi(u|x,\vec{w}) = \Gaus(u|\inv{\mat{K}}\vec{k},\inv{\mat{K}})~. 
\end{equation}
Such Gaussian policies are commonly employed in the continuous reinforcement learning setting, e.g., \cite{Bradtke94, Peters03}, and we emphasise that, as the state dependent part of the basis is largely unrestricted, this general class of basis sets does not seem unreasonably restrictive.

\subsubsection{Results}
\begin{figure}
\centering
  \subfloat[]{
    \input{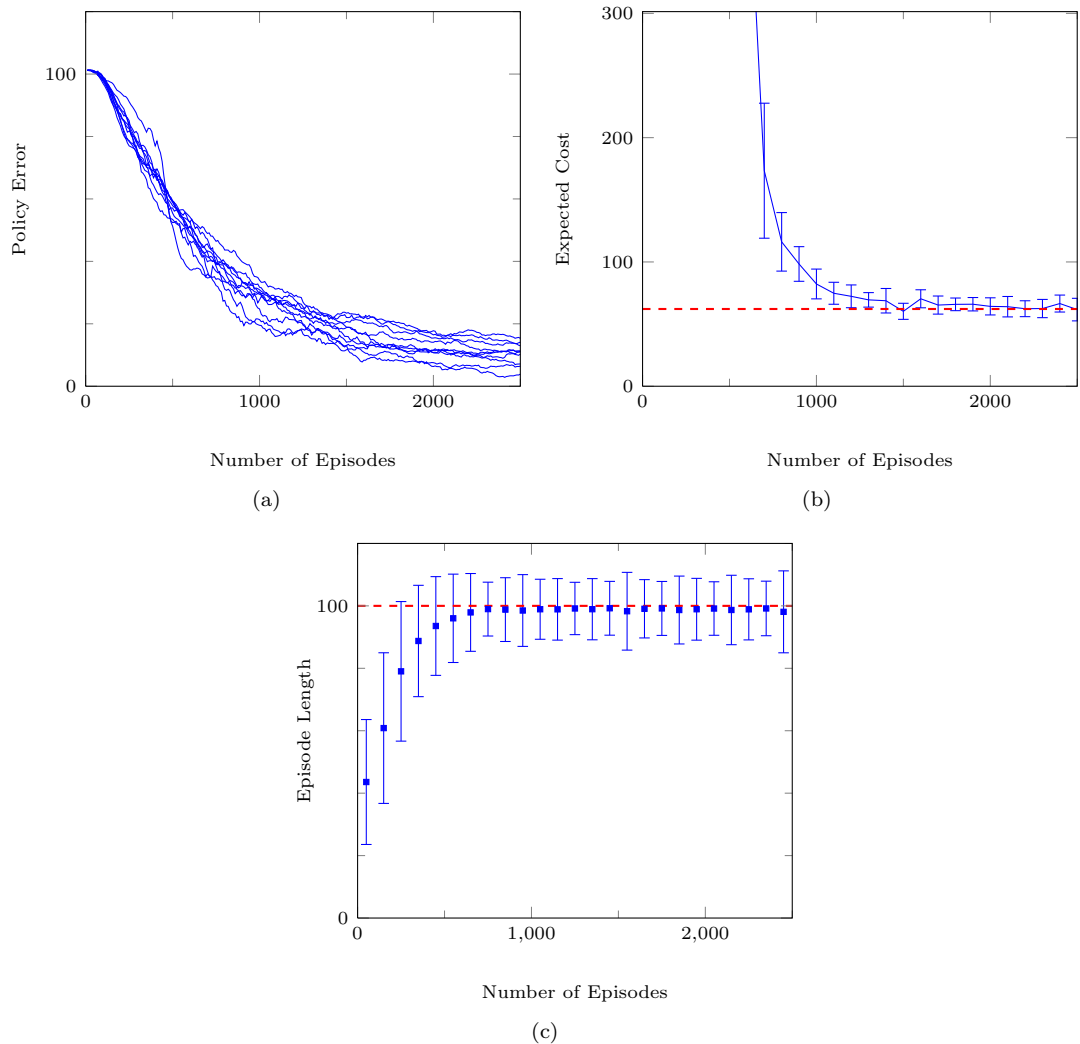}
  }
  \subfloat[]{
%
%
\begin{tikzpicture}
\begin{axis}[%
font={\scriptsize},
xlabel={Number of Episodes},
ylabel={Expected Cost},
scale only axis,
width =0.35\textwidth,
xmin=0, xmax=2500,
ymin=0, ymax=301.404,
axis on top,
xtick={0,1000,2000},
xticklabels={0,1000,2000},
minor x tick num=1,
every axis y label/.style=
{at={(-0.18,0.5)},rotate=90,anchor=center},
every axis x label/.style=
{at={(0.5,-0.2)},anchor=center}]

\addplot [
color=blue,
solid
]
plot [error bars/.cd, y dir = both, y explicit]
coordinates{ (647.29,316.628) +- (0,0) (700,173.412) +- (0,54.313) (800,116.175) +- (0,23.5439) (900,98.3666) +- (0,13.9323) (1000,82.2817) +- (0,11.9824) (1100,74.8039) +- (0,8.8407) (1200,72.3454) +- (0,9.16158) (1300,69.4876) +- (0,5.83782) (1400,68.7977) +- (0,9.85119) (1500,60.2808) +- (0,6.46423) (1600,70.3641) +- (0,7.25008) (1700,65.3127) +- (0,7.28123) (1800,65.891) +- (0,5.08608) (1900,65.9841) +- (0,5.38261) (2000,64.2594) +- (0,6.89418) (2100,64.0005) +- (0,8.21353) (2200,62.3777) +- (0,6.39207) (2300,62.5136) +- (0,7.34079) (2400,66.5328) +- (0,6.82006) (2500,61.6632) +- (0,9.07733)
};
\addplot [
thick,
color=red,
dashed
]
coordinates{ (0,62.1479) (2500,62.1479)
};

\end{axis}
\end{tikzpicture}
  }\\
  \subfloat[]{
%
%
\begin{tikzpicture}

\begin{axis}[%
font={\scriptsize},
xlabel={Number of Episodes},
ylabel={Episode Length},
scale only axis,
width=0.35\textwidth, 
xmin=0,  xmax=2500,
ymin=0, ymax=120,
axis on top,
xtick={0,1000,2000},
ytick={0,100},
minor x tick num=1,
minor y tick num=4,
every axis y label/.style=
{at={(-0.12,0.5)},rotate=90,anchor=center},
every axis x label/.style=
{at={(0.5,-0.2)},anchor=center}]

\addplot [
color=blue,
only marks,
mark=square*,
mark options={solid,scale=0.5}
]
plot [error bars/.cd, y dir = both, y explicit]
coordinates{ (50,43.575) +- (0,20.0093) (150,60.846) +- (0,24.1285) (250,79.023) +- (0,22.3541) (350,88.75) +- (0,17.8343) (450,93.551) +- (0,15.815) (550,96.01) +- (0,14.1532) (650,97.901) +- (0,12.4594) (750,98.966) +- (0,8.62364) (850,98.803) +- (0,10.2138) (950,98.513) +- (0,11.4951) (1050,98.913) +- (0,9.61352) (1150,98.882) +- (0,9.85409) (1250,99.166) +- (0,8.38396) (1350,98.929) +- (0,9.80061) (1450,99.228) +- (0,8.6186) (1550,98.279) +- (0,12.4531) (1650,99.066) +- (0,9.33814) (1750,99.179) +- (0,8.64151) (1850,98.669) +- (0,10.8707) (1950,98.915) +- (0,9.89736) (2050,99.147) +- (0,8.56906) (2150,98.681) +- (0,11.1214) (2250,98.886) +- (0,9.78912) (2350,99.172) +- (0,8.73927) (2450,98.088) +- (0,13.1237)
};

\addplot [
thick,
color=red,
dashed
]
coordinates{ (0,100) (2500,100)
};

\end{axis}
\end{tikzpicture}
  }
\caption{\label{fig:cartpole}Result for LS$\Psi$ for the cart on pole task. \textbf{(a)} Evolution of the error in the policy defined as the L$_2$ norm of the difference between learned and optimal gains for 10 random trails. \textbf{(b)} The evolution of the expected cost averaged over the 10 trials. The dashed line indicates the expected cost of the optimal policy. Error bars indicate standard deviation. \textbf{(c)} Average length of episodes in the 10 trials, averaged over blocks of 100 episodes.}
\end{figure}

We demonstrate the applicability of LS$\Psi$ on a pole on cart task \cite{SuttonBarto}, which has been repeatedly used as a benchmark in reinforcement learning \cite{Peters06, Peters07}.\exclude{The setup is illustrated in figure \ref{fig:cartonpole}.} The task consists of balancing a inverted pendulum mounted on a cart by exerting forces on the latter. The state space is given by $\vec{x} = (x,\dot{x}, \theta,\dot{\theta})$, with $x$ the position of the cart, $\theta$ the pendulums angular deviation from the upright postion and $\dot{x},\dot{\theta}$ their respective temporal derivatives. Following \cite{Peters03} we use a form of the dynamics linearised around the zero state. The approximate dynamics are given by $P(\yt|\xt,\ut) = \Gaus(\yt|\mat{A}\xt + \vec{b}\ut, \Sigma)$, where 
\begin{equation}
    \mat{A} = \left[\begin{array}{cccc}
                  1 & \tau & 0 & 0 \\ 
                  0 & 1    & 0 & 0 \\  
                  0 & 0    & 1 & \tau \\
                  0 & 0    &\nu\tau& 1  
              \end{array}\right]
              ~,\quad 
    \vec{b} = \left[\begin{array}{c}
                    0 \\
                    \tau \\
                    0 \\
                    \nu\tau/g
                    \end{array}
              \right]
\end{equation}
and $\tau = 1/60s$, $\nu=13.2s^{-2}$, $g=9.8ms^2$, $\Sigma=diag(0.001,0.001,0.001,0.001)$. The cost is given by $\C(x,u) = \trans{\vec{x}}\mat{Q}\vec{x} + u\mat{R}u$, with $\mat{Q} = diag(1.25, 1, 12, 0.25)$ and $\mat{R} = 0.01$, and was unlike in \cite{Peters03} not discounted, i.e, $\gamma =1$.

As the problem is LQG a set of polynomial basis functions is sufficiently rich to capture it, and  we applied LS$\Phi$ with bases 
\begin{equation*}
  \{u^2, ux, u\dot{x}, u\theta, u\dot{\theta}, x^2, x\dot{x}, x\theta, x\dot{\theta}, \dot{x}^2, \dot{x}\theta, \dot{x}\dot{\theta}, \theta^2, \theta\dot{\theta}, \dot{\theta}^2\} ~.    
\end{equation*}
This set is of the form required by \refeq{eq:approx_form}, if $w_1$ is negative. Although we employed no formal means to ensure this, empirically we found that if initialised to a negative value, $w_1$ would remain negative. Specifically we used the initialisation $\vec{w} = (-0.1, 0, \dots, 0)$ corresponding to an initial uniformed policy, i.e. a zero mean Gaussian with large variance. Using a random initialisation, whilst ensuring $w_1$ was negative did not affect the results significantly, although convergence times increased if the initial policy was far from the optimum and had a low variance. It is worth noting that the initial policy did not asymptotically stabilise the system, this is in contrast to \cite{Peters03, Peters07}\footnote{\cite{Peters03} did not require the initial policy to be stable, however a discounted cost was used and the initial policy was restricted to give $\gamma^{-2} > eig(\mat{A} - \vec{b}\vec{K})$ with $K$ the control gains, i.e. the policy had give rise to a well defined value function}.

We applied LS$\Psi$ following an episodic sampling procedure. Starting from a start state, drawn from $\Gaus(\vec{x}_0|0,\Sigma_0)$, with $\Sigma_0 = diag(0.5,0,0.1,0)$, a state, control \& cost trajectory was sampled according to the transition probability and cost function. The required controls were sampled according to the policy arising from the current $\vec{w}$, with a fixed baseline added to the variance. The latter proved necessary as otherwise the updates tended to become numerically unstable once the policy began to converge. A trajectory was terminated when it left the acceptable region given by 
\begin{equation}
  \label{eq:constraint}
  -\pi/6 \leq \theta \leq \pi/6 \quad \text{and} \quad -1.5m \leq x \leq 1.5m ~,
\end{equation}
as in \cite{Peters03}, or after 100 time steps. We updated $\vec{w}$ after every 10 episodes.

As the problem is LQG, the optimal policy is linear and can be computed directly. We can therefore asses the behaviour of \LSP\ directly, by measuring the error in the policy approximation during learning process. The results in figure \ref{fig:cartpole}(a), where we plot the policy error defined as the $\text{L}_2$ norm of the difference between the optimal gains and the \LSP\ estimate, demonstrate that \LSP\ can successfully find near optimal gains. As a, in the literature, more commonly reported metric of the quality of an RL algorithm is the expected cost under the policy it learns. In figure \ref{fig:cartpole}(b) we therefore plot the evolution of expected costs. Note that as the expected cost under certain policies for this problem is not finite, we plot a Monte Carlo estimate calculated from a set of 100 trajectories with 200 steps each\footnote{n.b. for the evaluation we did not apply constraints \refeq{eq:constraint}}. As a reference we also plot the expected cost under the optimal policy. This data confirms the results of the policy error analysis, i.e, that \LSP\ converges towards a near optimal policy. As an aside we note that these results are comparable in terms of the convergence time to the best performing methods in \cite{Peters06} were the same problem was used for evaluation. Unfortunately we were, so far, not able to directly reproduce these results in order to obtain a direct comparison. The similar convergence times are in particular surprising as \LSP\ started with a substantially worse initial policy. While \cite{Peters06} seem to have constrained the initial policies to be stable, the initial \LSP\ policy was unstable. This initial instability of the controlled system is illustrated in figure \ref{fig:cartpole}(c), where we plot the average length of the episodes used during learning. As can be seen the episodes under the initial policy are significantly shorter then the maximum length, indicating that the constraints in \refeq{eq:constraint} are frequently violated. However after about 600-700 episodes a stabilising policy is learnt.

\subsection{Conclusion}
The contribution of this section is a novel type of reinforcement learning algorithms, which we obtained by direct application of the theoretical insights of \autoref{sec:theory}. We were able to demonstrate that the proposed algorithm successfully solves classical problems. However we acknowledge that the performance compared to the state of the art remains to be investigated and we refrain from a full discussion until such data has been obtained.
  \bibliography{references}
  \newpage
\appendix
\appendixpage
\section{Supplementary proofs}
\label{appendix:lemmas}
\theoremstyle{plain}
\newtheorem*{lem}{Lemma}
\begin{lem}[\autoref{thm:minimizer} in \autoref{sec:theory:finite}]
Let $a,b,c$ be random variables with joint $P(a,b,c) = P(a)P(b\|a)P(c\|b,a)$ and $\set{P}$ the set of distributions over $a$, then
\begin{equation}
\label{local:eq:minimizer}
P(a)\exp\{\int_{b}P(b\|a) \log P(c=\hat{c}\|b)\} \varpropto \argmin_{q\in\set{P}} \KL{q(a)P(b\|a)}{P(a,b\|c = \hat{c})}
\end{equation}
and 
\begin{equation}
\label{local:eq:minimum}
\int_a P(a)\exp\{\int_{b}P(b\|a) \log P(c=\hat{c}\|b)\} = \min_{q\in\set{P}} \KL{q(a)P(b\|a)}{P(a,b\|c = \hat{c})}~.
\end{equation}
\end{lem}
\begin{proof}
We form the Lagrangian 
\begin{align}
\fun{L}
  & = \KL{q(a)P(b|a)}{P(a,b|c = \hat{c})} + \lambda\left[\int_a q(a) - 1\right]\\
  & \cong \int_{a,b} q(a)P(b|a)\log\frac{q(a)P(b|a)}{P(a)P(b|a)P(c = \hat{c}|b)} + \lambda\left[\int_a q(a) - 1\right]\\
  & = \int_{a} q(a)\log\frac{q(a)}{P(a)} - \int_{a,b} q(a)P(b|a)\log P(c = \hat{c}|b) ~,
\end{align}
where we use $\cong$ to indicate equality up to an additive constant. Setting the partial derivatives w.r.t. $q(a)$ to 0 gives
\begin{align}
0 = \log\frac{q(a)}{P(a)} + 1 - \int_{b} P(b|a)\log P(c = \hat{c}|b) + \lambda\\
  = \log\frac{q(a)}{\fun{Z}(\lambda)P(a)\exp\{\int_{b} P(b|a)\log P(c = \hat{c}|b)\}} ~,
\end{align}
where $\fun{Z}$ is a function of the lagrange multiplier. The result in \refeq{local:eq:minimizer} now directly follows and more specifically the minimizer is 
\begin{equation}
  q^*(a) = \frac{P(a)\exp\{\int_{b} P(b|a)\log P(c = \hat{c}|b)\}}{\int_a P(a)\exp\{\int_{b} P(b|a)\log P(c = \hat{c}|b)\}} ~.
\end{equation}
The result in \refeq{local:eq:minimum} can now easily be obtained by substituting $q^*$ into the $\opKL$ divergence.
\end{proof}
\exclude{
\begin{lemma}
  \label{thm:exp_ratio}
  Let $a > b$, then $\forall \eta > 0$, $\exists n \in \Integer$ s.t. $\exp\{na\} > \eta\exp\{nb\}$.
\end{lemma}
\begin{proof}
  Consider the ratio of $\exp\{na\}$ and $\exp\{nb\}$, which is $\exp\{n(a-b)\}$. By assumption $a - b > 0$, hence $\forall \eta > 0$, $\exists n \in \Integer$ s.t.  $\exp\{n(a-b)\} > \eta$ and as $\exp\{nb\} > 0$ the result follows.
\end{proof}
}

\end{document}